\documentclass{article}
\usepackage{ijcai24}

\usepackage{times}
\usepackage{soul}
\usepackage{url}
\usepackage[hidelinks]{hyperref}
\usepackage[utf8]{inputenc}
\usepackage[small]{caption}
\usepackage{graphicx}
\usepackage{amsmath}
\usepackage{amsthm}
\usepackage{booktabs}
\usepackage{algorithm}
\usepackage{algorithmic}
\usepackage[switch]{lineno}
\usepackage{amssymb}
\usepackage{multirow}
\usepackage{wrapfig}
\usepackage{marvosym}
\usepackage{makecell} 
\usepackage{colortbl}
\usepackage{color,xcolor}
\usepackage{subcaption}

\newtheorem{theorem}{Theorem}

\title{Rethinking Centered Kernel Alignment in Knowledge Distillation}

\author{
Zikai Zhou$^1$
\and
Yunhang Shen$^3$\and
Shitong Shao$^1$\and
Linrui Gong$^4$\and
Shaohui Lin$^{1,2}$\footnote{Corresponding author} \\
\affiliations
$^1$School of Computer Science and Technology, East China Normal University, Shanghai, China\\
$^2$Key Laboratory of Advanced Theory and Application in Statistics and Data Science, MOE, China\\
$^3$Youtu Lab, Tencent, Shanghai, China\\ $^4$Hunan University, China
\emails
\{choukai003,shenyunhang01,1090784053sst,wslgqq277a,shaohuilin007\}@gmail.com
}

\begin{document}

\maketitle

\begin{abstract}
\label{sec:abstract}
Knowledge distillation has emerged as a highly effective method for bridging the representation discrepancy between large-scale models and lightweight models. Prevalent approaches involve leveraging appropriate metrics to minimize the divergence or distance between the knowledge extracted from the teacher model and the knowledge learned by the student model. Centered Kernel Alignment (CKA) is widely used to measure representation similarity and has been applied in several knowledge distillation methods. However, these methods are complex and fail to uncover the essence of CKA, thus not answering the question of how to use CKA to achieve simple and effective distillation properly. This paper first provides a theoretical perspective to illustrate the effectiveness of CKA, which decouples CKA to the upper bound of Maximum Mean Discrepancy~(MMD) and a constant term. Drawing from this, we propose a novel Relation-Centered Kernel Alignment~(RCKA) framework, which practically establishes a connection between CKA and MMD. Furthermore, we dynamically customize the application of CKA based on the characteristics of each task, with less computational source yet comparable performance than the previous methods. The extensive experiments on the CIFAR-100, ImageNet-1k, and MS-COCO demonstrate that our method achieves state-of-the-art performance on almost all teacher-student pairs for image classification and object detection, validating the effectiveness of our approaches. Our code is available in \href{https://github.com/Klayand/PCKA}{\textcolor{purple}{https://github.com/Klayand/PCKA}}.
%

\end{abstract}
\section{Introduction}
\label{sec:intro}
 Tremendous efforts have been made in compressing large-scale models into lightweight models.
 Representative methods include network pruning~\cite{parameterpruning}, model quantization~\cite{wu_quantized_2016}, neural architecture search~\cite{NAS} and knowledge distillation~(KD)~\cite{vanillakd}.
 Among them, KD has recently emerged as one of the most flourishing topics due to its effectiveness~\cite{Liu_2021_ICCV,DIST,Gong2023AdaptiveHF,Shao2022TeachingWY} and wide applications~\cite{Chong2022MonoDistillLS,chen2023robust,Shao2023CatchUpDY}.
 Particularly, the core idea of KD is to transfer the acquired representations from a large-scale and high-performing model to a lightweight model by distilling the learning representations in a compact form, achieving precision and reliable knowledge transfer.
 Out of the consensus of researchers, there are two mainstream approaches to distilling knowledge from the teacher to the student.
 The first approach is the logit-based distillation, which aims to minimize the probabilistic prediction (response) scores between the teacher and student by leveraging appropriate metrics~\cite{DKD,vanillakd}.
 The other is feature-based distillation, which investigates the knowledge within intermediate representations to further boost the distillation performance~\cite{MGD,ICKD,ReviewKD,VID}.
 Among them, the design of metrics is essential in knowledge transfer and has been attractive to scholars.
 Kornblith \emph{at al}~\shortcite{CKA} proposes the Centered Kernel Alignment~(CKA) for the quantitative understanding of representations between neural networks.
 CKA not only focuses on model predictions but also emphasizes high-order feature representations within the models, providing a comprehensive and enriched knowledge transfer.


\begin{figure*}[t]
\includegraphics[width=1.2\textwidth]{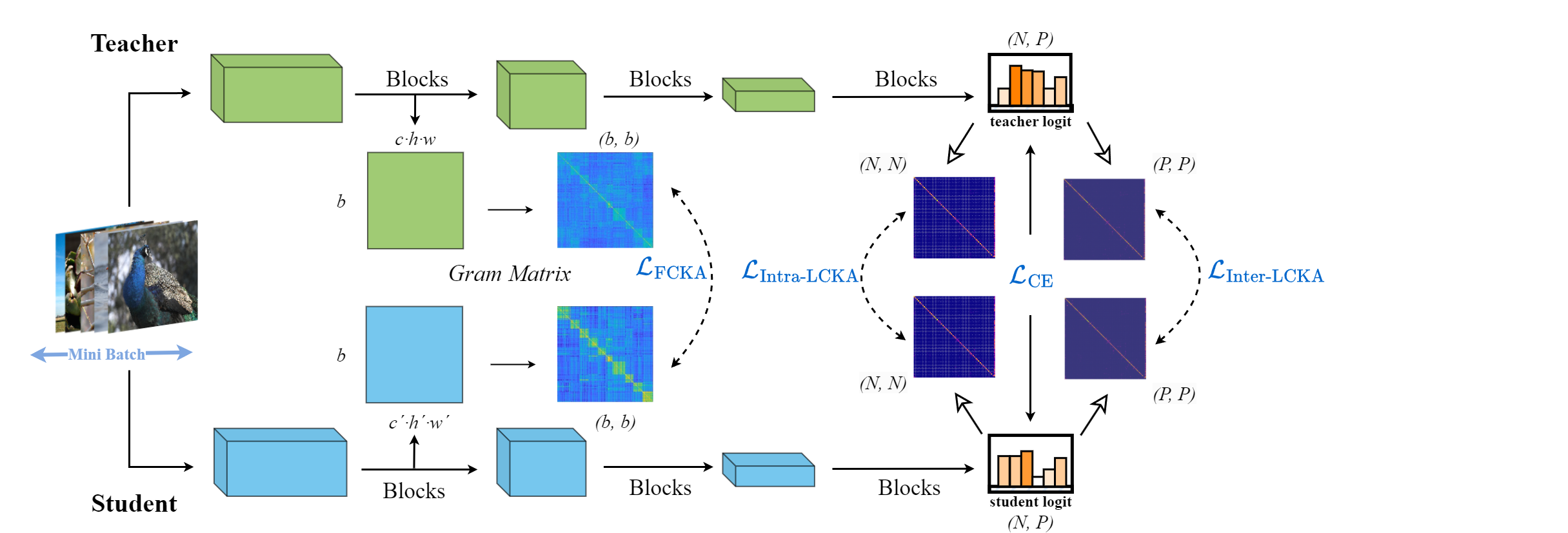}
\caption{
The overall framework of the proposed Relation-based Centered Kernel Alignment~(RCKA).
We first transform the feature map from the shape of $(B, C, HW)$ into $(B, CHW)$ and then compute the CKA similarity of feature maps between the teacher and the student.
Besides, we compute the inter-class and intra-class CKA similarity of logits between teacher and student.
Here, $N$ refers to the number of samples, and $P$ refers to the corresponding probability of class to which this sample belongs.}
\label{fig:cka}
\end{figure*}

Recent studies~\cite{DPK,bmvc} introduce CKA to quantitatively narrow the gap of learned representations between the teacher model and the student model.
Undeniably, these methods have achieved significant success.
However, their designs are excessively complex and need a large amount of computational resources, making it challenging to achieve fine-grained knowledge transfer and leading to low scalability.
Moreover, these methods fail to uncover the essence of CKA, lacking an in-depth analysis of CKA in knowledge distillation.
The reason why CKA is effective has not been explored.
Therefore, we focus on the theoretical analysis of CKA and rethink a more reasonable architecture design that ensures simplicity and effectiveness while generalizing well across various tasks.

In this paper, we provide a novel perspective to illustrate the effectiveness of CKA, where CKA is regarded as the upper bound of Maximum Mean Discrepancy~(MMD) with a constant term, specifically.
Drawing from this, we propose a Relation-Centered Kernel Alignment (RCKA) framework, which practically establishes a connection between CKA and MMD.
%
Besides, we dynamically customize the application of CKA on instance-level tasks, and introduce Patch-based Centered Kernel Alignment~(PCKA), with less computational source yet competitive performance when compared to previous methods.
Our method is directly applied not only to logit-based distillation but also to feature-based distillation, which exhibits superior scalability and expansion.
We utilize CKA to compute high-order representation information both between and within categories, which better motivates the alleviation of the performance gap between the teacher and student.
%
%
%

%
To validate the effectiveness of our approaches, we conduct extensive experiments on image classification (CIFAR-100~\cite{CIFAR} and ImageNet-1k~\cite{ILSVRC15}), and object detection (MS-COCO~\cite{COCO}) tasks.
As a result, our methods achieve state-of-the-art~(SOTA) performance in almost all quantitative comparison experiments with fair comparison.
Moreover, following our processing architecture, the performance of the previous distillation methods is further boosted in the object detection task.

Our contribution can be summarized as follows:
\begin{itemize}
\setlength{\itemsep}{0pt}
\setlength{\parsep}{0pt}
\setlength{\parskip}{0pt}
        \item[$\bullet$] We rethink CKA in knowledge distillation from a novel perspective, providing a theoretical reason for why CKA is effective in knowledge distillation.
        \item[$\bullet$]
        %
        We propose a Relation Centered Kernel Alignment (RCKA) framework to construct the relationship between CKA and MMD, with less computational source yet comparable performance than previous methods, which verifies our theoretical analysis correctly.
        %
        %
        %
        \item[$\bullet$]
        %
        We further dynamically customize the application of CKA for instance-level tasks and propose a Patch-based Centered Kernel Alignment~(PCKA) architecture for knowledge distillation in object detection, which further boosts the performance of previous distillation methods.
        \item[$\bullet$]
        We conduct plenty of ablation studies to verify the effectiveness of our method, which achieves SOTA performance on a range of vision tasks.
        Besides, we visualize the characteristic information of CKA and discover new patterns in it.
\end{itemize}

%
%
\section{Related Work}
\label{related_work}

Vanilla Knowledge Distillation~\cite{vanillakd} proposes aligning the output distributions of classifiers between the teacher and student by minimizing the KL-divergence, during training the emphasis on negative logits can be fine-tuned through a temperature coefficient, which serves as a form of normalization during the training process of a smaller student network. Tremendous efforts~\cite{SPKD,DIST,DPK,ATKD,RKD} have been made on how to design a good metric to align the distribution between the teacher and student.

Design an suitable alignment method for KD can start from two typical types: Drawing on representations, numerous methods have made significant strides by aligning the intermediate features~\cite{ATKD}, the samples' correlation matrices~\cite{SPKD}, and the output logits  between the teacher and student~\cite{DIST}.
From a mathematical standpoint, some measure theories are introduced to illustrate the similarity between the teacher and student, such as mutual information~\cite{VID}. Among these, Centered Kernel Alignment (CKA) is a valuable function for measuring similarity. It simultaneously considers various properties during similarity measures, such as invariance to orthogonal transformations. While the effectiveness of CKA in KD has been demonstrated in some works~\cite{DPK,bmvc}, the essence of CKA has not been thoroughly explored, and the unavoidable additional computational costs also limit its application prospect.

In this paper, we will revisit CKA in KD and provide a novel theoretical perspective to prove its effectiveness and analyze how it functions across various distillation settings.

\begin{figure*}[ht]
\centering
\includegraphics[width=1.0\textwidth]{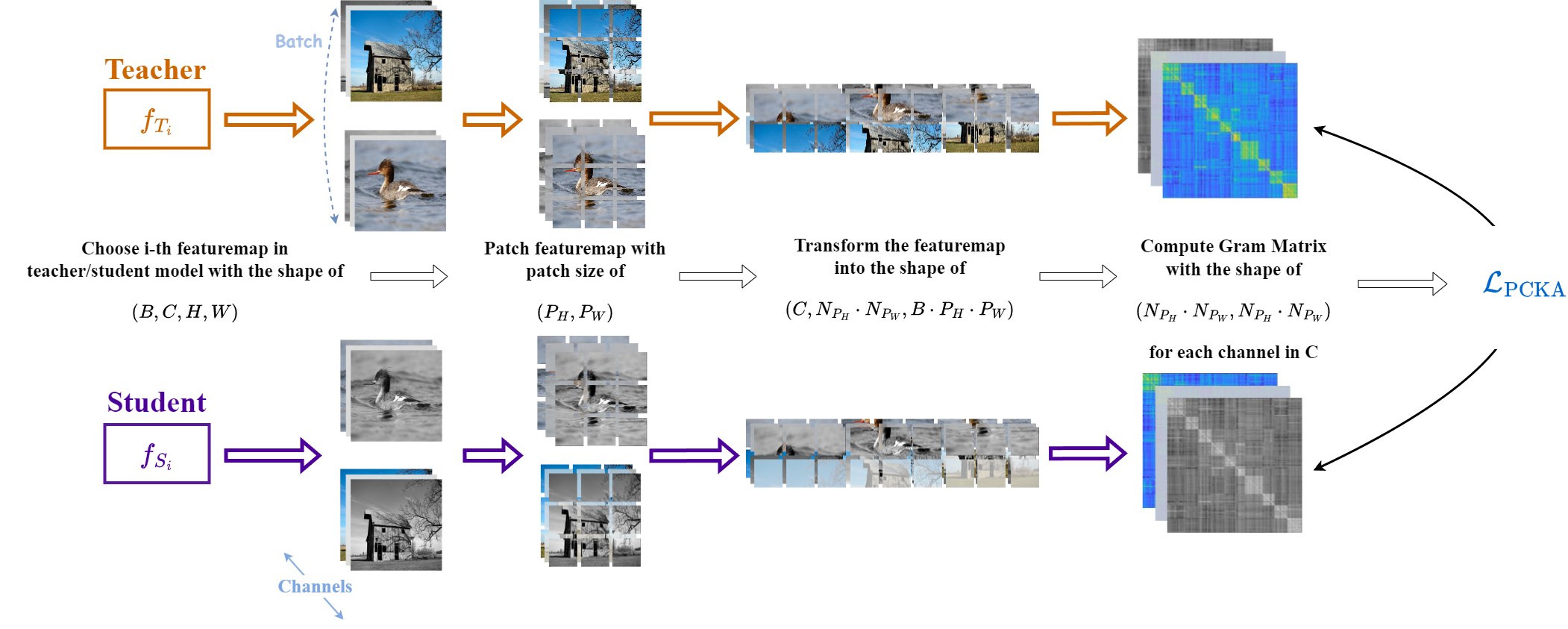}
\caption{The overall framework of PCKA. We dynamically customize the framework of proposed method based on the characteristics of object detection. In this framework, we first patch the featuremap of the teacher and student with the patchsize $(P_H, P_W)$, then transform the featuremap in order to get the gram matrix between each patch. Finally, we calculate the loss $L_{PCKA}$, and get average from dimension $C$. Here, $B, C, H,W$ refer to the batchsize, channels, height and width of the featuremap, respectively. $N_{P_H}, N_{P_W}$ denote the number of patches cutting along the height and width, respectively.}
\label{fig:det}
\end{figure*}

\section{Methodolgy}
\label{method}

\begin{table*}[t]
\center
\caption{
Results on the CIFAR-100 test set.
``Same'' and ``Different'' in the first row refer to whether the model architecture is the same for teachers and students. Combined distillation type means that this method transfers the knowledge both on the features and logits. Our methods surpass almost all algorithms with the same distillation type.
``RN'', ``WRN'', ``SN'', and ``MN''  denote ResNet, Wide ResNet, ShuffleNet, and MobileNet, respectively.
}
\begin{small}
\resizebox{1.\textwidth}{!}{%
\begin{tabular}{ll|cccccc|ccc}
\multicolumn{2}{c|}{Architecture} & \multicolumn{6}{c|}{Same} & \multicolumn{3}{c}{Different} \\\hline
\multirow{4}{*}{\begin{tabular}[c]{@{}c@{}}Distillation \\ Type\end{tabular}} & \multirow{2}{*}{Teacher}  & RN-110 & RN-110 & WRN-40-2 & WRN-40-2 & RN-32$\times$4 & VGG-13 & WRN-40-2 & RN-32$\times$4  & VGG-13\\
&      & 74.31     & 74.31     & 75.61      & 75.61      &  79.42    & 74.64 & 75.61 & 79.42 & 74.64  \\

& \multirow{2}{*}{Student}  & RN-20 & RN-32 & WRN-40-1  & WRN-16-2 & RN-8$\times$4 & VGG-8 & SN--V1 & SN--V1 & MN--V2 \\
& \space    & 69.06      & 71.14       & 71.98       & 73.26      & 72.50     & 70.36  &  70.50 & 70.50 & 64.60 \\ \hline
\multirow{12}{*}{Feature-based}
& FitNet~\shortcite{FitNet} &  68.99 & 71.06   & 72.24      & 73.58      & 73.50    & 71.02 & 73.73 & 73.59 &  64.14 \\
& ATKD~\shortcite{ATKD} & 70.22 & 70.55 & 72.77 & 74.08 & 73.44 & 71.43 & 73.32 & 72.73 & 59.40 \\
& SPKD~\shortcite{SPKD} & 70.04 & 72.69 & 72.43 & 73.83 & 72.94 & 72.68 & 74.52 & 73.48 & 66.30\\
& CCKD~\shortcite{CCKD} & 69.48 & 71.48 & 72.21 & 73.56 & 72.97 & 70.71 & 71.38 & 71.14 & 64.86 \\
& RKD~\shortcite{RKD} & 69.25  & 71.82 &  72.22 & 73.35 & 71.90 & 71.48 & 72.21 & 72.28 & 64.52 \\
& VID~\shortcite{VID} & 70.16 & 70.38 & 73.30 & 74.11 & 73.09 & 71.23 & 73.61 & 73.38 & 65.56 \\
& CRD~\shortcite{CRD}   & 71.46      & 73.48       & 74.14       & 75.48      & 75.51      & 73.94  & 76.05 & 75.11 & 69.73 \\
& OFD~\shortcite{COFD}   &  -   & 73.23    & 74.33   & 75.24 & 74.95 & 73.95  & 75.85 & 75.98 & 69.48\\
& ReviewKD~\shortcite{ReviewKD} &  - & 71.89 & 75.09 & 76.12 & 75.63 & 74.84 & 77.14 & 76.93 & 70.37 \\
& ICKD-C~\shortcite{ICKD} &  71.91 & 74.11 & 74.63 & 75.57 & 75.48 & 73.88 & 75.19 & 74.34 & 67.55 \\
& DPK~\shortcite{DPK} & \textbf{72.44} & \textbf{74.89} & 75.27 & 76.42 & - & 74.96 & 74.43 & 76.00 & 68.63 \\
& FCKA(ours) &  71.49 & 73.64 & 74.70 & 75.53 & 74.93 & 74.35 & 75.98 & 75.67 & 68.97 \\ \hline 
\multirow{4}{*}{Logit-based}       
& KD~\shortcite{vanillakd} & 70.67 & 73.08 & 73.54 & 74.92 & 73.33 & 72.98 & 74.83 & 74.07 & 67.37 \\
& DKD~\shortcite{DKD} & - & 74.11 & 74.81 & 76.24 & 76.32 & 74.68 & 76.70 & 76.45 & 69.71 \\
& DIST~\shortcite{DIST} & 69.94 & 73.55 & 74.42 & 75.29 & 75.79 & 73.74 & 75.23 & 75.23 & 68.48  \\
& IKL-KD~\shortcite{Cui2023DecoupledKD} & - & 74.26 & 74.98 & 76.45 & \textbf{76.59} & 74.88 & 77.19 & 76.64 & \textbf{70.40}\\
& NKD~\shortcite{Yang2023FromKD} & 71.26 & 73.79 & 75.23 & 76.37  & 76.35 & 74.86 & 76.59 & 76.90 & 70.22 \\
& LCKA(ours) & 70.87 & 73.64 & 74.63 & 75.78 & 75.12 & 74.35 & 76.12 & 76.43 & 69.37  \\ \hline
\multirow{2}{*}{Combined}
& SRRL~\shortcite{SRRL} & 71.51 & 73.80 & 74.75 & 75.96 & 75.92 & 74.40 & 76.61 & 75.66 & 69.14 \\
& RCKA(ours) & 72.26 & 74.31 & \textbf{75.34} & \textbf{76.51} & 76.11 & \textbf{74.97} & \textbf{77.21} & \textbf{76.97} & 70.12 \\
\end{tabular}}
\end{small}
\label{tab:cifar100result}
\end{table*}

In this section, we first revisit the paradigm of knowledge distillation and then introduce the formula of Centered Kernel Alignment~(CKA). Specifically, we derive the formula of the relationship between CKA and Maximum Mean Discrepancy~(MMD), where CKA can be decoupled as the upper bound of MMD with a constant term.
In light of the above deduction, we outline the methodology of our paper. We apply the proposed methods in image classification and object detection, dynamically customizing CKA for each task.

\subsection{The Paradigm of Knowledge Distillation}
The existing KD methods can be categorized into two groups.
Particularly, the logits-based KD methods narrow the gap between the teacher and student models by aligning the soft targets between them, which is formulated as following loss term:
\begin{align}
\label{logits}
    L_\mathrm{logits} = \mathsf{D}_\mathrm{logits}(\mathsf{T}_s(\sigma(z_s; \tau)), \mathsf{T}_t(\sigma(z_t; \tau))),
\end{align}
where $z_s$ and $z_t$ are the logits from students and teachers, respectively.
And $\sigma(\cdot)$ is the softmax function that produces the category probabilities from the logits, and $\tau$ is a non-negative temperature hyper-parameter to scale the smoothness of the predictive distribution.
Specifically, we have $\sigma_i(z;\tau) = \operatorname{softmax}(\exp(z_i/\tau))$.
$\mathsf{D}_\mathrm{logits}$ is a loss function to capture the discrepancy distributions, \emph{e.g.} Kullback-Leibler divergence.
And $\mathsf{T}_s$ and $\mathsf{T}_t$ denote the transformation functions in students and teachers, respectively, which usually refer to the identity mapping in Vanilla KD~\cite{vanillakd}.
Similarly, the feature-based KD methods, which aim to mimic the feature representations between teachers and students, are also represented as a loss item:
\begin{align}
\label{feature}
    L_\mathrm{feat} = D_\mathrm{feat}(\mathsf{T}_s(\mathbf{F}_s), \mathsf{T}_t(\mathbf{F}_t)),
\end{align}
where $\mathbf{F}_s$ and $\mathbf{F}_t$ denote feature maps from students and teachers, respectively.
Transformation modules $\mathsf{T}_s$ and $\mathsf{T}_t$ align the dimensions of $\mathbf{F}_s$ and $\mathbf{F}_t$.
$D_\mathrm{feat}$ computes the distance between two feature maps, such as $\ell_1$- or $\ell_2$ norm.
Therefore, the KD methods can be represented by a generic paradigm.
The final loss is the weighted sum of the cross-entropy loss $L_{ce}$, the logits distillation loss, and the feature distillation loss:
\begin{align}
\label{kd}
    L = L_{ce} + \alpha L_\mathrm{logits} + \beta L_\mathrm{feat},
\end{align}
where $\alpha$ and $\beta$ are hyper-parameters controlling the trade-off between these three losses.

\subsection{Distilling with the Upper Bound}
Centered Kernel Alignment~(CKA) has been proposed as a robust way to measure representational similarity between neural networks. We first prove that CKA measures the cosine similarity of the gram matrix between teachers and students. 
%
%
\begin{theorem}
    [Proof in \href{https://arxiv.org/abs/2401.11824}{Appendix}~\ref{appendix:proof:cosine}]
    \label{theorem:cosine}
    Let $X$ and $Y$ be $N \times P$ matrices.
    The CKA similarity \( \|Y^\top X\|_F^2 \) is equivalent to the cosine similarity of $XX^\top$ and $YY^\top$, which denote the gram matrix of $X$ and $Y$, respectively.
    In other words, 
    \begin{equation*}
    \begin{aligned}
        S_\mathrm{CKA}(X, Y) & = \frac{\|Y^\top X\|_F^2}{\|X^TX\|_F\|Y^TY\|_F} \\
        & = \frac{\mathrm{vec}(X X^\top)^\top \mathrm{vec}(Y Y^\top)}{\|\mathrm{vec}(X X^\top)\|_2 \|\mathrm{vec}(Y Y^\top) \|_2},
    \end{aligned}
    \end{equation*}
    where $\mathrm{vec}$ operator represents reshaping the matrix to a vector.
\end{theorem}

We then derive the formula of the relationship between CKA and MMD, where CKA can be regarded as the upper bound of MMD with a constant term. 

\begin{theorem}
    [Proof in \href{https://arxiv.org/abs/2401.11824}{Appendix}~\ref{appendix:proof:mmd}]
    \label{theorem:MMD}
    Maximizing CKA similarity is equivalent to minimizing the upper bound of MMD distance:
    \begin{equation*}
        \begin{aligned}
        &\frac{\|Y^\top X\|_F^2}{\|X^TX\|_F\|Y^TY\|_F} = - N \mathbb{E}_{i,j} [\left<x_i, x_j\right>- \left<y_i, y_j\right>]^2 +2 \\
        &\leq - N \left(\mathbb{E}_{i,j}[\left<x_i, x_j\right>]- \mathbb{E}_{i,j}[\left<y_i, y_j\right>]\right)^2 +2,
        \end{aligned}
    \end{equation*}
    where the inequality is given by Jesen's inequality.
\end{theorem}

According to Jesen's inequality, CKA can be decoupled as the upper bound of MMD with a constant term. The first term corresponds to minimizing the upper bound of MMD distance with the RKHS kernel. In contrast, the latter constant term acts as a weight regularizer, enhancing the influence of MMD, where it promotes the similarity between features of the same batch, not only instances in the same class but also in different classes.
On one hand, optimizing the upper bound of MMD, which has additional stronger constraints, allows it to converge to the optimal solution more quickly and stably.
On the other hand, the latter term serves as a weight scaling mechanism, effectively avoiding the challenges of optimization caused by excessively small MMD values, which result in small gradients. 

According to the deduction, we successfully transformed our optimization objective from maximizing CKA to minimizing the upper bound of MMD, which makes our method more intuitive and concise. Building upon these findings, we propose our methods, which are more effective than previous methods.

\subsection{Relation-based Centered Kernel Alignment}
As illustrated in Fig.~\ref{fig:cka}, we propose a Relation Centered Kernel Alignment~(RCKA) framework in image classification.
In this framework, we leverage CKA as a loss function to ensure that the centered similarity matrix is distilled rather than forcing the student to mimic the teacher`s similarity matrix with a different scale.
This is very important because a model`s discriminative capability is dependent on the distribution of its features rather than its scale, which is inconsequential for class separation~\cite{Nguyen2020DoWA,Orhan2017SkipCE}.

Assume we have a large-scale teacher model $t$ and a lightweight student model $s$.
The activation map from layer $l$ of the teacher is denoted as $F^{(l)}_t \in \mathbb{R}^{b \times c \times h \times w}$, whereas the activation map of layer $l^{\prime}$ of the student is denoted as $F^{(l^{\prime})}_s \in \mathbb{R}^{b \times c^{\prime} \times h^{\prime} \times w^{\prime}}$.
%
$c$, $h$, and $w$ denote the channel, height, and width of the teacher, whereas $c^{\prime}, h^{\prime}$ and $w^{\prime}$ denote that of the student.
The mini-batch size is denoted by $b$.
The logits of the teacher and student are denoted as $z_t \in \mathbb{R}^{N\times P}$ and $z_s \in \mathbb{R}^{N\times P^{\prime}}$, where $N$ and $P$~(or $P^{\prime}$) refer to the number of samples and the corresponding probability for which class this sample belongs to.
%
Therefore, the formula of our method, similar to Eqn.~\ref{kd}, is represented as:
\begin{align}
\label{rcka:equation}
    L_\mathrm{RCKA} & = L_{CE} + \alpha L_\mathrm{FCKA} \\
    & + \beta (L_\mathrm{Intra-LCKA} + L_\mathrm{Inter-LCKA})
\end{align}
where $\alpha$ and $\beta$ are hyper-parameters controlling the trade-off between the features loss $L_\mathrm{FCKA}$ and logits loss $(L_\mathrm{Intra-LCKA} + L_\mathrm{Inter-LCKA})$. 
The $L_\mathrm{FCKA}$, $L_\mathrm{Intra-LCKA}$ and $L_\mathrm{Inter-LCKA}$ are represented as:
\begin{align}
\label{rcka:each}
    &L_\mathrm{FCKA} = S_\mathrm{CKA}(\mathsf{T}(\mathbf{F}_t), \mathsf{T}(\mathbf{F}_s)), \\
    &L_\mathrm{Intra-LCKA} = S_\mathrm{CKA}(z_t, z_s), \\
    &L_\mathrm{Inter-LCKA} = S_\mathrm{CKA}(z_t^\mathsf{T}, z_s^\mathsf{T}),
\end{align}
where $\mathsf{T}$ in Eqn.~6 refers to the transformation module $\mathbb{R}^{b \times c \times h \times w} \rightarrow \mathbb{R}^{b \times chw}$.
%

%
Compared with the previous methods, our method has superior scalability and expansion and can be directly applied to both feature and logits distillation.
We calculate the gram matrix to collect high-order inter-class and intra-class representations, encouraging the student to learn more useful knowledge.
Also, we provide the relationship between CKA and MMD in \href{https://arxiv.org/abs/2401.11824}{Appendix}~\ref{appendix:proof:mmd} to better demonstrate the theoretical support of our method.
%


%

Because the value of CKA ranges from $\lbrack 0,1 \rbrack$, at the beginning of the training process, $L_{CE}$ plays a more important role than all CKA losses to drive the optimization of the student, which helps the student avoid matching extremely complex representations.

\begin{table*}[htp]
\caption{Results on the ImageNet validation set. We use ResNet-34 and ResNet-50 released by Torchvision~\protect\cite{marcel2010torchvision} as our teacher's pre-training weight.}
\renewcommand\arraystretch{0.95}
\setlength\tabcolsep{0.6mm}
	\center
 \footnotesize
\resizebox{1.\textwidth}{!}{%
\begin{tabular}{cc|c|cc|ccccc|cccc|cc}
\multicolumn{2}{c|}{Architecture} & &
\multicolumn{2}{c|}{Accuracy}  & \multicolumn{5}{c|}{Feature-based} & \multicolumn{4}{c|}{Logit-based} & \multicolumn{2}{c}{Combined} \\\hline
Teacher & Student &  & Teacher & Student & OFD & CRD & ReviewKD & ICKD-C & MGD~\shortcite{MGD} & KD & RKD & DKD & DIST & SRRL & Ours \\\hline
\multirow{2}{*}{ResNet-34} & \multirow{2}{*}{ResNet-18} & Top-1 & 73.31 & 69.76 & 71.08 & 71.17 & 71.61  & 72.19 & 71.80 & 70.66 & 70.34 & 71.70 & 72.07 & 71.73 & \textbf{72.34} \\
		~ & & Top-5 & 91.42 & 89.08 & 90.07 & 90.13 & 90.51 & 90.72 & 90.40 & 89.88 & 90.37 & 90.41 & 90.42 & 90.60 & \textbf{90.68} \\
		\hline
	    \multirow{2}{*}{ResNet-50} & \multirow{2}{*}{MobileNet-V1} & Top-1 & 76.16 & 70.13 & 71.25 & 71.37 & 72.56 & - & 72.59 & 70.68 & - & 72.05 & \textbf{73.24} & 72.49 & 72.79\\
		~ & & Top-5 & 92.86 & 89.49 & 90.34 & 90.41 & 91.00 & - & 90.74 & 90.30 & - & 91.05 & \textbf{91.12} & 90.92 & 91.01\\ \end{tabular}}
\label{tab:imagenet_result}
\end{table*}
\begin{table*}[ht]
\caption{Results on the COCO validation set (T$\rightarrow$S refers to the distillation from T to S). Here, the content in brackets to the right of ``Ours'' refers to the methods applied in the distillation process. In addition, CM RCNN-X101 stands for Cascade Mask RCNN-X101.}
\resizebox{1.\textwidth}{!}{
\begin{tabular}{l|cccccc|cccccccl|cccccc}
 \multirow{1}{*}{T$\rightarrow$S} & \multicolumn{6}{c|}{CM RCNN-X101~\shortcite{Cai2017CascadeRD}$\rightarrow$Faster RCNN-R50~\shortcite{Ren2015FasterRT}} & \multicolumn{6}{c}{RetinaNet-X101$\rightarrow$RetinaNet-R50~\shortcite{Lin2017FocalLF}} & &  \multirow{1}{*}{T$\rightarrow$S}
 &\multicolumn{6}{c}{FCOS-R101$\rightarrow$FCOS-R50~\shortcite{Tian2019FCOSFC}} \\
 
 Type & \multicolumn{6}{c|}{\textit{Two-stage detectors}} & \multicolumn{6}{c}{\textit{One-stage detectors}} & & Type & \multicolumn{6}{c}{\textit{Anchor-free detectors}}\\
Method & AP & AP$_{50}$ & AP$_{75}$ & AP$_{S}$ & AP$_{M}$ & AP$_{L}$ & AP & AP$_{50}$ & AP$_{75}$ & AP$_{S}$ & AP$_{M}$ & AP$_{L}$ & & Method & AP & AP$_{50}$ & AP$_{75}$ & AP$_{S}$ & AP$_{M}$ & AP$_{L}$ \\\cline{0-12}\cline{15-21}

Teacher & 45.6 & 64.1 & 49.7 & 26.2 & 49.6 & 60.0 & 41.0 & 60.9 & 44.0 & 23.9 & 45.2 & 54.0 & &Teacher & 40.8 & 60.0 & 44.0 & 24.2 & 44.3 & 52.4 \\
Student & 38.4 & 59.0 & 42.0 & 21.5 & 42.1 & 50.3 & 37.4 & 56.7 & 39.6 & 20.0 & 40.7 & 49.7& & Student & 38.5 & 57.7 & 41.0 & 21.9 & 42.8 & 48.6 \\\cline{0-12}\cline{15-21}
KD~\shortcite{vanillakd} & 39.7 & 61.2 & 43.0 & 23.2 & 43.3 & 51.7 & 37.2 & 56.5 & 39.3 & 20.4 & 40.4 & 49.5 & & KD~\shortcite{vanillakd} & 39.9 & 58.4 & 42.8&23.6& 44.0 & 51.1\\
COFD~\shortcite{COFD} & 38.9 & 60.1 & 42.6 & 21.8 & 42.7 & 50.7 & 37.8 & 58.3 & 41.1 & 21.6 & 41.2 & 48.3 & & FitNet~\shortcite{FitNet} & 39.9 & 58.6 & 43.1 & 23.1 & 43.4 & 52.2 \\
FKD~\shortcite{FKD} & 41.5 & 62.2 & 45.1 & 23.5 & 45.0 & 55.3 & 39.6 & 58.8 & 42.1 & 22.7 & 43.3 & 52.5 & & GID~\shortcite{GID} &42.0 & 60.4 & \textbf{45.5} & 25.6 & 45.8 & 54.2 \\
DIST~\shortcite{DIST} & 40.4 & 61.7 & 43.8 & 23.9 & 44.6 & 52.6 & 39.8 & 59.5 & 42.5 & 22.0 & 43.7 & 53.0 & & FRS~\shortcite{FRS} &40.9 & 60.3 & 43.6 & 25.7 & 45.2 & 51.2  \\
DIST+mimic~\shortcite{DIST} & 41.8 & 62.4 & 45.6 & 23.4 & 46.1 & 55.0 & 40.1 & 59.4 & 43.0 & 23.2 & 44.0 & 53.6 & & FGD~\shortcite{FGD} & \textbf{42.1} & - & - & \textbf{27.0}& \textbf{46.0}& \textbf{54.6} \\
Ours & 41.4 & 62.1 & 45.2 & 23.5 & 45.6 & 54.9 & 40.3 & 59.9 & 43.0 & 23.3 & 44.2 & 54.9 && Ours & 39.8 & 59.0 & 42.4 & 22.2 & 43.6 & 52.5\\
Ours + mimic&\textbf{42.4} &\textbf{63.3}&\textbf{46.1}&\textbf{24.3}&\textbf{46.7}&\textbf{56.1}&\textbf{40.7}&\textbf{60.4}&\textbf{43.4}&\textbf{23.9}&\textbf{44.7}&\textbf{55.1}&&Ours + mimic & 40.7 & \textbf{60.5} & 43.1 & 23.4 & 44.8 & 53.1 \\
\end{tabular}}
\label{tab:detectionresult}
\end{table*}

\subsection{Patch-based Centered Kernel Alignment}
%

%
In this subsection, we further adapt the proposed RCKA to instance-level tasks such as object detection.
However, directly applying RCKA to instance-level tasks may deteriorate performance, as the above tasks are usually trained with a small size of mini-batches (\textit{e.g.} $2$ or $4$ per GPU), causing the failure of the gram matrix to collect enough knowledge.
Besides, increasing the mini-batch size requires a significant amount of computational resources, making it infeasible in practice.
Thus, we dynamically customize our RCKA method for object detection.
Recent works~\cite{CWD,COFD} find that distilling the representations of intermediate layers is more effective than distilling the logits in object detection.
Therefore, we adjust our method to only target intermediate layers.
Still, we follow our core idea in the classification task, which calculates the similarities between different instances by using CKA.
So, we divide the image feature maps into several patches and compute the similarities between different patches.
Our redesigned method is illustrated in Fig.~\ref{fig:det}.
In this framework, we first patch the feature maps of the teacher and student with a patch size of $(P_H, P_W)$, then transform the feature maps to get the gram matrix between each patch.
Finally, we calculate the loss $L_\mathrm{PCKA}$ and get the average from dimension $C$.
Here, $N_{P_H}$ and $N_{P_W}$ denote the number of patches cutting along the height and width, respectively.
Therefore, the Patch-based CKA loss is represented as:
\begin{align}
\label{pcka}
    L_\mathrm{PCKA} = \gamma S_\mathrm{CKA}(N^s_{P_H} \cdot N^s_{P_W}, N^t_{P_H} \cdot N^t_{P_W}),
\end{align}
where $N^s_{P_H} \cdot N^s_{P_W}$ and $N^t_{P_H}, N^t_{P_W}$ are denoted as the number of the student patches and the teacher patches, respectively.
Usually, $N^s_{P_H} \cdot N^s_{P_W} = N^t_{P_H}, N^t_{P_W}$. $\gamma$ refers to the loss weight factor.

\section{Experiments}

We conduct extensive experiments on \emph{image classification} and \emph{object detection} benchmarks.
The image classification datasets include CIFAR-100~\cite{CIFAR} and ImageNet-1k~\cite{ILSVRC15} and the object detection dataset includes MS-COCO~\cite{COCO}.
Moreover, we present various ablations and analyses for the proposed methods.
More details about these datasets are in \href{https://arxiv.org/abs/2401.11824}{Appendix}~\ref{apd:dataset}.
We apply a batch size of $128$ and an initial learning rate of $0.1$ for the SGD optimizer on CIFAR-100.
And we follow the settings in~\cite{DIST} for the ResNet34-ResNet18 pair and the ResNet50-MobileNet pair on ImageNet-1k.
The settings of other classification and detection tasks are in \href{https://arxiv.org/abs/2401.11824}{Appendix}~\ref{apd:hsettings}.
Our code will be publicly available for reproducibility.

\subsection{Image Classification}
\paragraph{Classification on CIFAR-100.}
We compare state-of-the-art (SOTA) feature-based and logit-based distillation algorithms on $\textbf{9}$ student-teacher pairs.
%
Among them, $6$ pairs have the same structure for teachers and students, and the rest of them have different architectures.
The results are presented in Tab.~\ref{tab:cifar100result}.
Our proposed method outperforms all other algorithms on $4$ student-teacher pairs and achieves comparable performance on the rest of them, meanwhile requiring extremely less computational resources and time consumption than the SOTA methods DPK~\cite{DPK} and ReviewKD~\cite{ReviewKD}.
The comparisons of computational cost are in \href{https://arxiv.org/abs/2401.11824}{Appendix}~\ref{apd:time}.

\paragraph{Classification on ImageNet-1k.}
We also conduct experiments on the large-scale ImageNet to evaluate our methods.
Our RCKA achieves comparable results with other algorithms, even outperforms them, as shown in Tab.~\ref{tab:imagenet_result}.
We find that with the increasing of categories and instances, it is more challenging for the student to mimic the high-order distribution of the teacher. Moreover, in \href{https://arxiv.org/abs/2401.11824}{Appendix} Tab.~\ref{tab:deit_imagenet}, we explore the feature distillation for ViT-based models on ImageNet-1k. It is noted that our method outperforms other methods, which means that our method has good scalability and good performance.

\begin{table}[t]
\caption{Results on the COCO validation set (T$\rightarrow$S refers to the distillation from T to S). Here, the content in brackets to the right of ``Ours'' refers to the methods applied in the distillation process. In addition, “Inside GT Box” means we use the GT boxes with the same stride on the FPN layers as the feature imitation regions. “Main Region” means we imitate the features within the main distillation region.}
\resizebox{1.\linewidth}{!}{
\begin{tabular}{l|cccccc}
 \multirow{1}{*}{T$\rightarrow$S} & \multicolumn{6}{c}{GFL-R101$\rightarrow$GFL-R50 ~\shortcite{GFL}}  \\ \hline
Method & AP & AP$_{50}$ & AP$_{75}$ & AP$_{S}$ & AP$_{M}$ & AP$_{L}$ \\ \hline
Teacher & 44.9 & 63.1 & 49.0 & 28.0 & 49.1 & 57.2  \\
Student & 40.2 & 58.4 & 43.3 & 23.3 & 44.0 & 52.2  \\ \hline
FT~\shortcite{FitNet} & 40.7 & 58.6 & 44.0 & 23.7 & 44.4 & 53.2 \\
Inside GT Box & 40.7 & 58.6 & 44.2 & 23.1 & 44.5 & 53.5 \\
DeFeat & 40.8 & 58.6 & 44.2 & 24.3 & 44.6 & 53.7 \\
Main Region & 41.1 & 58.7 & 44.4 & 24.1 & 44.6 & 53.6 \\
FGFI~\shortcite{FGFI} & 41.1 & 58.8 & 44.8 & 23.3 & 45.4 & 53.1 \\
FGD~\shortcite{FGD} & 41.3 & 58.8 & 44.8 & 24.5 & 45.6 & 53.0 \\
GID~\shortcite{GID} & 41.5 & 59.6 & 45.2 & 24.3 & 45.7 & 53.6 \\
SKD~\shortcite{SKD} & 42.3 & 60.2 & 45.9 & 24.4 & 46.7 & 55.6 \\
Our & \textbf{42.8} & \textbf{61.2} & \textbf{46.3} & \textbf{24.8} & \textbf{47.1} & \textbf{55.4} \\
\end{tabular}}
\label{tab:other_det_2}
\end{table}

\subsection{Object Detection}
\paragraph{Detection on MS-COCO.}
Comparison experiments are run on three kinds of different detectors, \emph{i.e.}, tow-stage detectors, one-stage detectors, and anchor-free detectors.
As shown in Tab.~\ref{tab:detectionresult}, PCKA outperforms the precious methods almost on all three kinds of metrics, by aligning the high-order patch-wise presentations.
We believe that aligning feature maps of the student and teacher in low-order could also improve the performance of PCKA, driven by mimicking low-order representations in the early stage and then learning high-order and complex representations gradually.
Thus, we follow~\cite{DIST} by adding auxiliary mimic loss, \emph{i.e.}, translating the student feature maps from the teacher feature map by a convolution layer and supervising them utilizing $\mathcal{L}_{MSE}$, to the detection distillation task.
Ultimately, we conclude from Tab.~\ref{tab:detectionresult} that PCKA-based mimic loss achieves the best performance on Cascade RCNN-X101-Cascade RCNN-R50 and RetinaNet-X101-RetinaNet-R50 pairs.
We also conduct experiments on the other four architectures, as shown in Tab.~\ref{tab:other_det_2} and Tab.~\ref{tab:other_det_1}. These results further validate the effectiveness of our proposed method.

\subsection{Ablations and Visualizations}
We conduct ablation studies in three aspects:
\textbf{(a)} the effect of hyper-parameters.
\textbf{(b)} effectiveness of the proposed modules.
\textbf{(c)} unexplored phenomenon during training.

\paragraph{Ablation studies on hyperparameters.}
As shown in Tab.~\ref{tab:batchsize}, Tab.~\ref{tab:weight} and Tab.~\ref{tab:distillation_layer} in \href{https://arxiv.org/abs/2401.11824}{Appendix}, we conduct the ablation studies on the size of mini-batch, loss scaling factor $\gamma$ on $L_{PCKA}$ and the number of intermediate layers for distilling.
We find the local optima values are the mini-batch size $12$, loss scaling factor $10$, and $3$ layers of distillation.

\paragraph{The upper bound of MMD.}
In Theorem~\ref{theorem:MMD}, we derive the relationship between CKA and MMD, where CKA is the upper bound of MMD with a constant term. To validate this, We conduct the experiment, which is shown in Tab.~\ref{tab:strong}, we notice that CKA, which is the upper bound of MMD, has additional stronger constraints. Because of this, CKA converges to the optimal solution more quickly and stably, compared with MMD.

\paragraph{The dimension to average.}
In PCKA framework, we cut the activations of the teacher and student in the shape of $(C, N_{P_H} \cdot N_{P_W}, B \cdot P_H \cdot P_W)$.
We also carry out the experiments of averaging on different dimensions, shown in Tab.~\ref{tab:dim_avg}.
We find averaging on channel dimension is the optima.

\paragraph{Patch distillation.}
We explore the effectiveness of cutting activations into patches.
As shown in Tab.~\ref{tab:with_patch}, several standard distillation methods~\cite{vanillakd,ATKD,DIST} all perform well with patch cutting, validating the effectiveness of cutting patches.
With the smaller representation distribution in patches, it is easier to align the teacher and student.
Thus, the proposed PCKA architecture amazingly boosts the previous methods.
%

\begin{table}[ht]
\caption{The ablation study on the COCO validation set (T$\rightarrow$S refers to the distillation from T to S). Here, we explore which dimension we should choose to get better results. "Mix-up" means the 1st distilling layer uses Batch avg. method, the 2nd distilling layer uses Spatial avg. method and the final distilling layer uses Channel avg. method}
\resizebox{1.\linewidth}{!}{
\begin{tabular}{l|cccccc}
 \multirow{1}{*}{T$\rightarrow$S} & \multicolumn{6}{c}{RetinaNet-X101$\rightarrow$RetinaNet-R50}  \\ \hline
Method & AP & AP$_{50}$ & AP$_{75}$ & AP$_{S}$ & AP$_{M}$ & AP$_{L}$ \\ \hline
Teacher & 41.0 & 60.9 & 44.0 & 23.9 & 45.2 & 54.0  \\
Student & 37.4 & 56.7 & 39.6 & 20.0 & 40.7 & 49.7  \\ \hline
Batch avg. & 38.5 & 57.9 & 40.8 & 20.7 & 41.5 & 52.5 \\
Spatial avg. & 39.3 & 58.7 & 41.9 & 21.4 & 41.3 & 50.9 \\
Mix-up avg. & 38.2 & 58.1 & 40.4 & 21.3 & 41.9 & 50.9 \\
Channel avg. & \textbf{40.3} & \textbf{59.9} & \textbf{43.0} & \textbf{23.3} & \textbf{44.2} & \textbf{54.9} \\
\end{tabular}}
\label{tab:dim_avg}
\end{table}

\paragraph{Visualize the CKA value.}
We present some visualizations to show that our method does bridge the teacher-student gap in logit-level.
In particular, we visualize the logit similarity for $6$ teacher-student pairs in \href{https://arxiv.org/abs/2401.11824}{Appendix}~\ref{apd:cka_value}.
We find that our method significantly improves the logit-similarity.
%

\paragraph{Visualize the training process.}
We further visualize the training process of different detectors and the patch effect on the RetinaNet-X101-RetinaNet-R50 pair.
The results are in Tab.~\ref{fig:avg} and Tab.~\ref{fig:training} in the \href{https://arxiv.org/abs/2401.11824}{Appendix}.

\begin{table}[ht]
\caption{Experiments on the upper bound of MMD. We derive the formula that CKA is the upper bound of MMD with a constant term. From these experiments, we can prove that optimizing the upper bound of MMD can better improve the performance, compared with MMD.}
\resizebox{1.\linewidth}{!}{
\begin{tabular}{l|cccccc}
 \multirow{1}{*}{T$\rightarrow$S} & \multicolumn{6}{c}{RetinaNet-X101$\rightarrow$RetinaNet-R50}  \\ \hline
Method & AP & AP$_{50}$ & AP$_{75}$ & AP$_{S}$ & AP$_{M}$ & AP$_{L}$ \\ \hline
Teacher & 41.0 & 60.9 & 44.0 & 23.9 & 45.2 & 54.0  \\
Student & 37.4 & 56.7 & 39.6 & 20.0 & 40.7 & 49.7  \\ \hline
MMD w/ patch & 38.5 & 57.7 & 40.9 & 22.2 & 42.8 & 51.3 \\
PCKA & \textbf{40.3} & \textbf{59.9} & \textbf{43.0} & \textbf{23.3} & \textbf{44.2} & \textbf{54.9} \\
\end{tabular}}
\label{tab:strong}
\end{table}

\paragraph{Visualize the inference outputs.}
We first visualize the confusion matrix of the proposed method in Fig.~\ref{fig:confusion}, and then visualize the annotated images of training with/without patches averaging different dimensions in Fig.~\ref{fig:vis}, respectively. These figures reveal that our method can collect the similarities between different classes, and also show the effectiveness of our method on the object detection task.

\section{Discussion}
\paragraph{PCKA in image classification.}
We apply PCKA to the classification task, and it also outperforms well on the methods with the same distillation type, as shown in Tab.~\ref{tab:cifar100patch}.
However, PCKA performs very badly on the teacher-student pairs with different architectures.
As cutting activations of different architectures contain more dissimilar and harmful representations, bringing difficulty in transferring knowledge to the student.

\paragraph{Average on the channel, boosting the performance.}
The results in Tab.~\ref{tab:with_patch} reveal an interesting phenomenon, where the performance of the previous distillation methods is boosted by averaging the loss on channel dimension after the activations are cut into patches.
Instead of directly matching the whole representation distribution in the activations, cutting patches makes the alignment between the teacher and student easier with a smaller representation distribution in patches.
Besides, cutting into patches follows the idea in the framework of classification proposed by us, thus PCKA calculates the inter-class similarities and intra-class similarities in patches.
Moreover, due to the superiority of cosine similarity over distance-based losses~\cite{Boudiaf2020AUM} and high-order distribution representations collected by the gram matrix, PCKA outperforms DIST and AT.

\paragraph{Positional information loss.}
In PCKA, we cut the activation of the teacher and student into patches, and then flatten them into a vector.
Although this operation damages the original positional information, performance does not deteriorate.
We suppose that CKA ensures the focus of the optimization is on the shape of the distribution, rather than the raw values in the Gram matrix, which is vital because a model`s discriminative capability is dependent on the distribution of its features rather than its scale.
Besides, at the beginning, the effect brought by PCKA is smaller, compared with CE loss.
Therefore, CE loss motivates the optimization of the student model steadily, and starting from a certain moment, PCKA drives the student model to align complex and high-order representations, improving the generalization ability.

\begin{table}[t]
\caption{Ablation study of distillation methods with(w/) or without(w/o) patch on the COCO validation set (T$\rightarrow$S refers to the distillation from T to S). Here, we surprisingly notice that previous distillation method performance can be improved by image patching.}
\resizebox{1.\linewidth}{!}{
\begin{tabular}{l|cccccc}
 \multirow{1}{*}{T$\rightarrow$S} & \multicolumn{6}{c}{RetinaNet-X101$\rightarrow$RetinaNet-R50}  \\ \hline
Method & AP & AP$_{50}$ & AP$_{75}$ & AP$_{S}$ & AP$_{M}$ & AP$_{L}$ \\ \hline
Teacher & 41.0 & 60.9 & 44.0 & 23.9 & 45.2 & 54.0  \\
Student & 37.4 & 56.7 & 39.6 & 20.0 & 40.7 & 49.7  \\ \hline
KD  & 37.2 & 56.5 & 39.3 & 20.4 & 40.4 & 49.5 \\
KD w/ patch & 39.3 & 58.7 & 41.9 & 21.4 & 41.3 & 50.9 \\
AT & 34.4 & 52.3 & 36.4 & 17.7 & 37.2 & 47.8 \\
AT w/ patch & 37.4 & 56.6 & 39.9 & 20.8 & 40.6 & 49.8 \\
DIST & 39.8 & 59.5 & 42.5 & 22.0 & 43.7 & 53.0 \\
DIST w/ patch & 40.2 & 59.6 & \textbf{43.2} & 22.7 & \textbf{44.8} & 53.9 \\
PCKA w/o patch & 36.4 & 55.8 & 38.7 & 20.6 & 39.8 & 48.7 \\
PCKA(ours)  & \textbf{40.3} & \textbf{59.9} & 43.0 & \textbf{23.3} & 44.2 & \textbf{54.9} \\
\end{tabular}}
\label{tab:with_patch}
\end{table}
\section{Conclusion}
In this paper, we provide a novel theoretical perspective of CKA in knowledge distillation, which can be simplified as the upper bound of MMD with a constant term.
Besides, we dynamically customize the application of CKA based on the characteristics of each task, with less computational source yet comparable performance than previous methods.
Furthermore, we propose a novel processing architecture for knowledge distillation in object detection task, which can further boost the performance of previous distillation methods.
Our experimental results, including both qualitative and quantitative ones, demonstrate the effectiveness of our methods.
In future research, we will further explore the relationship between all similarity metric-based distillation methods, and explore the theoretical reason why averaging on the channel dimension with patches can boost the performance of previous methods.

\section*{Acknowledgements}
This work is supported by the National Natural Science Foundation of China (NO. 62102151), Shanghai Sailing Program (21YF1411200), CCF-Tencent Rhino-Bird Open Research Fund, the Open Research Fund of Key Laboratory of Advanced Theory and Application in Statistics and Data Science, Ministry of Education (KLATASDS2305), the Fundamental Research Funds for the Central Universities.


\small{
\bibliographystyle{named}
\bibliography{ijcai24}
}

\clearpage
\clearpage
\appendix

\begin{table*}[htp]
\caption{Results on the ImageNet validation set. $*$ indicates the teacher is pre-trained on ImageNet-21K. We evaluate our method on DeiT architectures.}
\renewcommand\arraystretch{0.95}
\setlength\tabcolsep{0.6mm}
	\center
 \footnotesize
 \begin{small}
\centering
\resizebox{.8\textwidth}{!}{%
\begin{tabular}{cc|c|cc|ccc}
\multicolumn{2}{c|}{Architecture} & &
\multicolumn{2}{c|}{Accuracy}  & \multicolumn{3}{c}{Methods}  \\\hline
Teacher & Student &  & Teacher & Student & KD & VIT-KD~\shortcite{Yang2022ViTKDPG} & VIT-KD+RCKA \\\hline
\multirow{2}{*}{DeiT \uppercase\expandafter{\romannumeral3}-Small*~\shortcite{DeiT}} & \multirow{2}{*}{DeiT-Tiny} & Top-1 & 82.76 & 74.42 & 76.01 & 76.06 & \textbf{77.23} \\
		~ & & Top-5 & - & 92.29 & 93.26 & 93.16 & \textbf{93.67}  \\
		\hline
	    \multirow{2}{*}{DeiT \uppercase\expandafter{\romannumeral3}-Base*} & \multirow{2}{*}{DeiT-Small} & Top-1 & 85.48 & 80.55 & 82.52 & 81.95 & \textbf{83.12} \\
		~ & & Top-5 & - & 95.12 & 96.30 & 95.64 & \textbf{96.41} \\ \end{tabular}}
\end{small}
\label{tab:deit_imagenet}
\end{table*}

\section{Datasets}
\label{apd:dataset}
\paragraph{CIFAR-100.} Dataset CIFAR-100~\cite{CIFAR} is the subsets of the tiny image dataset and consists of 60,000 images with the size 32$\times$32. Specifically, the training set contains 50,000 images, and the testing set contains 10,000 images. 
\paragraph{ImageNet-1k.} Dataset ImageNet-1k~\cite{ILSVRC15}, also commonly referred to as ILSVRC 2012, has 1000 classes, and the benchmark is trained using the training set and tested using the validation set. Its training and validation sets contain 1281,167 and 50,000 images, respectively.
\paragraph{MS-COCO.} Dataset MS-COCO~\cite{COCO} is a large-scale object detection dataset. The benchmark is the same as ImageNet-1k, using the training set for training and the validation set for testing. The training/validation split was changed from 83K/41K to 118K/5K in 2017. Researchers commonly apply the 2017 version for experiments.
\paragraph{Cityscapes.} Dataset Cityscapes~\cite{cordts2016cityscapes} is a new large-scale dataset for semantic segmentation. It provides 5,000 images that have been meticulously annotated, with 2,975 images for training and 500 images for validation, where 30 common classes are provided and 19 classes are used for evaluation and testing. Each image is 2048$\times$1024 in size.

\begin{table*}[ht]
\caption{Results on the COCO validation set (T$\rightarrow$S refers to the distillation from T to S). Here, the content in brackets to the right of ``Ours'' refers to the methods applied in the distillation process.}
\resizebox{1.\textwidth}{!}{
\begin{tabular}{l|cccccc|cccccccl|cccccc}
 \multirow{1}{*}{T$\rightarrow$S} & \multicolumn{6}{c|}{Faster RCNN-R101$\rightarrow$Faster RCNN-R50} & \multicolumn{6}{c}{RetinaNet-R101$\rightarrow$RetinaNet-R50} & &  \multirow{1}{*}{T$\rightarrow$S}
 &\multicolumn{6}{c}{RepPoints-X101$\rightarrow$RepPoints-R50~\shortcite{RepPointsPS}} \\ \hline
Method & AP & AP$_{50}$ & AP$_{75}$ & AP$_{S}$ & AP$_{M}$ & AP$_{L}$ & AP & AP$_{50}$ & AP$_{75}$ & AP$_{S}$ & AP$_{M}$ & AP$_{L}$ & & Method & AP & AP$_{50}$ & AP$_{75}$ & AP$_{S}$ & AP$_{M}$ & AP$_{L}$ \\\cline{0-12}\cline{15-21}
Teacher & 39.9 & 60.1 & 43.3 & 23.5 & 44.2 & 51.5 & 38.9 & 58.0 & 41.5 & 21.0 & 42.8 & 52.4 & &Teacher & 44.2 & 65.5 & 47.8 & 26.2 & 48.4 & 58.5 \\
Student & 38.4 & 59.0 & 42.0 & 21.5 & 42.1 & 50.3 & 37.4 & 56.7 & 39.6 & 20.0 & 40.7 & 49.7& & Student & 38.6 & 59.6 & 41.6 & 22.5 & 42.2 & 50.4 \\\cline{0-12}\cline{15-21}
FGFI~\shortcite{FGFI} & 39.3 & 59.8 & 42.9 & 22.5 & 42.3 & 52.2 & 38.6 & 58.7 & 41.3 & 21.4 & 42.5 & 51.5 & & FKD~\shortcite{FKD} & 40.6 & - & - &23.4& 44.6 & 55.3\\
GID~\shortcite{GID} & 40.2 & 60.7 & 43.8 & 22.7 & 44.0 & 53.2 & 39.1 & 59.0 & 42.3 & 22.8 & 43.1 & 52.3 & & Our & \textbf{41.0} & \textbf{62.2} & \textbf{44.0} & \textbf{23.6} & \textbf{44.6} & \textbf{55.6} \\
FGD~\shortcite{FGD} & 40.4 & - & - & 22.8 & 44.5 & 53.5 & 39.6 & - & - & 22.9 & 43.7 & 53.6 & \\
Our & 40.1 & \textbf{60.7} & \textbf{43.8} & 22.7 & 44.1 & 52.6 & 38.6 & 57.8 & 41.3 & 20.7 & 42.3 & 51.9 &  \\
\end{tabular}}
\label{tab:other_det_1}
\end{table*}

\section{Hyperparameter Settings}
\label{apd:hsettings}
\paragraph{Classification.} For the classification experiments on CIFAR-100, the batch size is 128, the total number of epochs is 240, and the learning rate is initialized to 0.1 and scheduled by ALRS~\cite{chen2022bootstrap}. In addition, we employ an SGD optimizer for training and set the weight decay and momentum as 5e-4 and 0.9, respectively. For the classification experiments on ImageNet-1k (ResNet34-ResNet18 pair and ResNet50-MobileNet pair), the total batch size is 512, the total number of epochs is 240, the batch size in every GPU is 128, the number of GPUs is 4 and the learning rate is initialized to 0.1 and scheduled by ALRS. Besides, we employ an SGD optimizer for training and set the weight decay and momentum as 1e-4 and 0.9, respectively. The loss weight scaling factors $\alpha, \beta$ are both 5.

\paragraph{Detection.} For the detection experiments on MS-COCO, we utilize mmdetection~\cite{mmdetection} and mmrazor~\cite{2021mmrazor} for both training and testing. Following~\cite{CWD,RKD,Zhou2023AFNAF}, we use the same standard training strategies on the Cascade RCNN-X101-Faster RCNN-R50 and RetinaNet-X101-RetinaNet-R50 pairs. To be specific, the total batch size is 12, the total number of epochs is 24, the batch size in every GPU is 3, the number of GPUs is 4 and the learning rate is divided by 10 at 16 and 22 epochs. The initial learning rate is set as 0.02 and 0.01 on Cascade RCNN-X101-Faster RCNN-R50 and RetinaNet-X101-RetinaNet-R50 pairs, respectively. Besides, the setting on the FCOS-R101-FCOS-R50 pair is following~\cite{FGD}. Compared with the RetinaNet-X101-RetinaNet-R50 pair, the only difference is we apply a warm-up learning rate on the FCOS-R101-FCOS-R50 pair. Furthermore, the settings on GFL-R101-GFL-R50, RepPoints-X101-RepPoints-R5, RetinaNet-R101-RetinaNet-R50 and Faster RCNN-R101-Faster RCNN-R50 pairs are the same with the setting on RetinaNet-X101-RetinaNet-R50 pair.

\paragraph{Segmentation.} For the segmentation experiments on Cityscapes, we apply mmsegmentation~\cite{mmseg2020} and mmrazor~\cite{2021mmrazor} for distillation. we follow the setting in~\cite{2021mmrazor}. In specific, the total batch size is 16, the total number of iterations is 80,000, the batch size in every GPU is 2, the number of GPUs is 8 and the learning rate is 0.01. We make the learning rate decay to 0.9 in each iteration and constrain the minimum learning rate in training to be 1e-4. And we utilized a SGD optimizer for training and set the weight decay and momentum as 5e-4 and 0.9, respectively.

\section{Derivations}

\subsection{Revisit the formula of CKA}
\label{appendix:proof:cosine}
We prove CKA measures the cosine similarity between teacher`s gram matrix and student gram matrix.
\begin{proof}
    \begin{equation*}
        \begin{aligned}
            &\frac{\|Y^\top X\|_F^2}{\|X^TX\|_F\|Y^TY\|_F} = \frac{\mathrm{tr}((Y^\top X)^\top Y^\top X)}{\|X^TX\|_F\|Y^TY\|_F}  \\
            = &\frac{\mathrm{tr}(X^\top Y Y^\top X)}{\|X^TX\|_F\|Y^TY\|_F} 
            = \frac{\mathrm{tr}(X X^\top Y Y^\top )}{\|X^TX\|_F\|Y^TY\|_F}  \\
            =& \frac{\mathrm{vec}(X X^\top)^\top \mathrm{vec}(Y Y^\top)}{\|\mathrm{vec}(X X^\top)\|_2 \|\mathrm{vec}(Y Y^\top) \|_2}.
        \end{aligned}
    \end{equation*}
\end{proof}

\subsection{Connection between CKA and MMD}
\label{appendix:proof:mmd}
In Theorem \ref{theorem:cosine}, we already prove the relationship between CKA distance and cosine similarity. Denote the i-th row of matrix $\frac{X}{\sqrt{||XX^T||_F}}$ as $x_i$ and i-th row of matrix $\frac{Y}{\sqrt{||YY^T||_F}}$ as $y_i$, we can get:
\begin{align*}
    &\frac{\|Y^\top X\|_F^2}{\|X^TX\|_F\|Y^TY\|_F} \\
    = &\frac{\mathrm{vec}(X X^\top)^\top \mathrm{vec}(Y Y^\top)}{\|\mathrm{vec}(X X^\top)\|_2 \|\mathrm{vec}(Y Y^\top) \|_2} \\
    = &- \| \frac{\mathrm{vec}(X X^\top) }{\|\mathrm{vec}(X X^\top)\|_2 } - \frac{ \mathrm{vec}(Y Y^\top)}{ \|\mathrm{vec}(Y Y^\top) \|_2 } \|_2^2 \\
    + & \|\frac{\mathrm{vec}(X X^\top) }{\|\mathrm{vec}(X X^\top)\|_2 }\|_2^2 + \| \frac{ \mathrm{vec}(Y Y^\top)}{ \|\mathrm{vec}(Y Y^\top) \|_2}\|_2^2 \\
    = &- \| \frac{\mathrm{vec}(X X^\top) }{\|\mathrm{vec}(X X^\top)\|_2 } - \frac{ \mathrm{vec}(Y Y^\top)}{ \|\mathrm{vec}(Y Y^\top) \|_2 } \|_2^2 + 2 \\
    = &- \left( \sum_{i,j} [\left<x_i, x_j\right>- \left<y_i, y_j\right>]^2\right)
    + 2\\
    = &  - N \mathbb{E}_{i,j} [\left<x_i, x_j\right>- \left<y_i, y_j\right>]^2 +2 \\
    \leq& - N \left(\mathbb{E}_{i,j}[\left<x_i, x_j\right>]- \mathbb{E}_{i,j}[\left<y_i, y_j\right>]\right)^2 +2 \\
\end{align*}

Consequently, maximizing the Centered Kernel Alignment (CKA) similarity is tantamount to minimizing the upper bound of the Maximum Mean Discrepancy (MMD) distance.

\section{More comparisons with recent KD methods in detection.} We add new methods published recently for fair comparison in Tab.~\ref{tab:detection_result}. Our method outperforms TST and DRKD by $1.9\%$ and $0.8\%$ AP, respectively, when distilling CM RCNN-X101 $\rightarrow$ Faster RCNN-R5.
\begin{table}[ht]
\small
\caption{\small Comparison with recent KD methods in object detection.}
\resizebox{.5\textwidth}{!}{
\begin{tabular}{l|cccccc|cccccccl|cccccc}
 \multirow{1}{*}{T$\rightarrow$S} & \multicolumn{6}{c|}{CM RCNN-X101$\rightarrow$Faster RCNN-R50} & \multicolumn{6}{c}{RetinaNet-X101$\rightarrow$RetinaNet-R50}\\
Method & AP & AP$_{50}$ & AP$_{75}$ & AP$_{S}$ & AP$_{M}$ & AP$_{L}$ & AP & AP$_{50}$ & AP$_{75}$ & AP$_{S}$ & AP$_{M}$ & AP$_{L}$ \\\cline{0-12}

Teacher & 45.6 & 64.1 & 49.7 & 26.2 & 49.6 & 60.0 & 41.0 & 60.9 & 44.0 & 23.9 & 45.2 & 54.0 \\
Student & 38.4 & 59.0 & 42.0 & 21.5 & 42.1 & 50.3 & 37.4 & 56.7 & 39.6 & 20.0 & 40.7 & 49.7 \\\cline{0-12}
TST~\cite{Shao2022TeachingWY} & 40.5 & 62.4 & 44.1 & 24.0 & 44.6 & 52.1 & 39.9 & 59.6& 42.8 & 23.3 & 43.8 & 53.3 \\
DRKD~\cite{Ni2023DualRK} & 41.6 & 62.4 & 45.3 & 24.2 & 45.3 & 55.3 & 40.3 & 59.7 & 42.9 & 23.4 & 44.2 & 53.4 \\

Ours & 41.4 & 62.1 & 45.2 & 23.5 & 45.6 & 54.9 & 40.3 & 59.9 & 43.0 & 23.3 & 44.2 & 54.9 \\
Ours + mimic&\textbf{42.4} &\textbf{63.3}&\textbf{46.1}&\textbf{24.3}&\textbf{46.7}&\textbf{56.1}&\textbf{40.7}&\textbf{60.4}&\textbf{43.4}&\textbf{23.9}&\textbf{44.7}&\textbf{55.1}\\
\end{tabular}}
\label{tab:detection_result}
\end{table}

\section{Segmentation on Cityscapes.}
As shown in Table~\ref{tab:seg}, we extend our method to the segmentation task. Our method outperforms SKDS, SKD, and IFVD by $2.01\%$, $0.63\%$, and $0.17\%$ mIoU by distilling PSPNet-R101 $\rightarrow$ PSPNet-R18~\cite{pspnet}, respectively. The results reveal the capability of our method on the segmentation task.
\begin{table}[!h]
 \caption{Results on the Cityscapes validation set.}
    \renewcommand\arraystretch{1.1}
	\setlength\tabcolsep{3mm}
	\vspace{-0.1in}\centering
\scriptsize
\setlength{\tabcolsep}{2.8mm}{
\resizebox{.5\textwidth}{!}{
	\begin{tabular}{l|cccccc}
        
        Method & T: PSPNet-R101 & S: PSPNet-R18 & SKDS & SKD & IFVD & Ours\\ \hline  
        mIoU (\%) & 78.55 & 70.09 & 72.70 & 74.08 & 74.54 & \textbf{74.71}\\ 
	\end{tabular}}}
    \label{tab:seg}
\end{table}

\section{Quantification the computational cost}
\label{apd:time}

\paragraph{The average running time comparison.}In Tab.~\ref{run_time}, we show a superior average training time cost than other methods, illustrating the lightweight of our proposed method.
\begin{table}[ht]
    \centering
    \caption{Average throught (batches / second) of training ResNet-18 student with ResNet-34 teacher on ImageNet. The speed is tested based on our implementations on 8 NVIDIA V100 GPUs.}
    \resizebox{0.9\linewidth}{!}{
    \begin{tabular}{lllllllll}
    \hline
        Method & \textbf{KD} & \textbf{RKD} & \textbf{SRRL} & \textbf{CRD} & \textbf{DIST} & \textbf{RCKA} \\ \hline
        Throughout & 14.28 & 11.11 & 12.98 & 8.33 & 14.19 & 11.31
    \end{tabular}}
    \label{run_time}
\end{table}

\paragraph{The transformation block cost comparison.} We quantify the computational overhead of the transform module in Tab.~\ref{tab:cost}, highlighting that our method eliminates the usage of the transformation module, further reducing FLOPs and parameters.
\begin{table}[!ht]
\centering
\renewcommand\arraystretch{1.0}
\small
\caption{Computational cost quantification.}
\setlength{\tabcolsep}{0.6mm}{
\resizebox{.8\linewidth}{!}{
\begin{tabular}{lcccc}
\toprule
        \textbf{Methods} & \textbf{Teacher Transform} & \textbf{Student Transform} & \textbf{\#params} & \textbf{FLOPs} \\ \midrule
        Ours & None & None & 0 & 0.0G \\ \midrule
        AT & Attention & Attention & 9.3M & 1.7G \\ \midrule
        OFD & Margin ReLU & 1x1 conv & 13.3M & 2.5G \\ \midrule
        ReviewKD & 1x1 conv & 1x1 conv & 24.2M & 2.8G \\ \midrule
        DPK & 1x1 conv & 1x1 conv & 17.2M & 2.4G \\ \midrule
        DPK & None & MLP-Decoder & 18.3M & 2.5G \\ \bottomrule
    \end{tabular}}}
\label{tab:cost}
\end{table}

\begin{figure}[ht]
\includegraphics[width=1\linewidth]{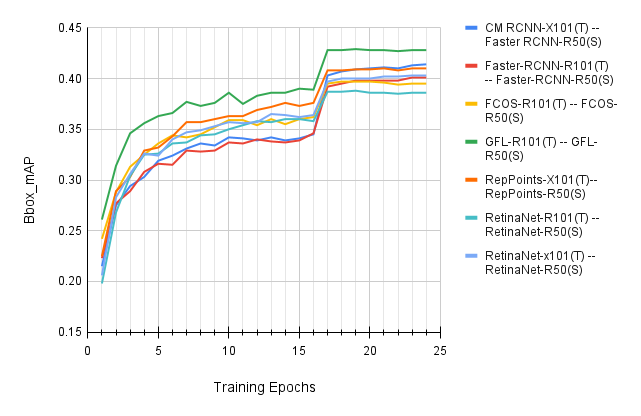}
\caption{Training process visualization of all experimented detectors.}
\label{fig:training}
\end{figure}

\begin{table}[!ht]
\vspace{-10pt}
\caption{Ablation study of loss weight $\gamma$.}
\resizebox{1.\linewidth}{!}{
\begin{tabular}{l|cccccc}
 \multirow{1}{*}{T$\rightarrow$S} & \multicolumn{6}{c}{RetinaNet-X101$\rightarrow$RetinaNet-R50}  \\ \hline
Loss Weight & AP & AP$_{50}$ & AP$_{75}$ & AP$_{S}$ & AP$_{M}$ & AP$_{L}$ \\ \hline
Teacher & 41.0 & 60.9 & 44.0 & 23.9 & 45.2 & 54.0  \\
Student & 37.4 & 56.7 & 39.6 & 20.0 & 40.7 & 49.7  \\ \hline
 $\gamma = 5$  & 39.8 & 59.4 & 42.5 & 22.3 & 43.5 & 53.9 \\
 $\gamma = 10$ & \textbf{40.3} & \textbf{59.9} & \textbf{43.0} & \textbf{23.3} & \textbf{44.2} & \textbf{54.9} \\
 $\gamma = 15$ & 38.5 & 58.0 & 41.0 & 20.7 & 41.8 & 51.3 \\
\end{tabular}}
\label{tab:weight}
\end{table}

\begin{table}[ht]
\centering
\caption{Results on the CIFAR-100 test set. We apply PCKA to the classification task, and it also outforms some of the previous methods.}
\begin{small}
\resizebox{0.5\textwidth}{!}{%
\begin{tabular}{ll|cccc}
\hline
\multirow{4}{*}{\begin{tabular}[c]{@{}c@{}}Distillation \\ Type\end{tabular}} & \multirow{2}{*}{Teacher}  & ResNet110 & ResNet110 & WRN-40-2~\shortcite{Zagoruyko2016WRN} & VGG13  \\
&      & 74.31     & 74.31     & 75.61     & 74.64  \\
& \multirow{2}{*}{Student}  & ResNet20~\shortcite{ResNet} & ResNet32  & WRN-16-2 & VGG8  \\
& \space    & 69.06      & 71.14     & 73.26      & 72.50  \\ \hline
\multirow{4}{*}{}
& FitNet~\shortcite{FitNet} &  68.99 & 71.06  &73.58 & 71.02 \\
& ATKD~\shortcite{ATKD} & 70.22 & 70.55 & 74.08  & 71.43  \\
& SPKD~\shortcite{SPKD} & 70.04 & 72.69 & 73.83 & 72.68 \\
& CCKD~\shortcite{CCKD} & 69.48 & 71.48 & 73.56 & 70.71 \\
& RKD~\shortcite{RKD} & 69.25  & 71.82 & 73.35 & 
71.48 \\
& VID~\shortcite{VID} & 70.16 & 70.38 & 74.11 
& 71.23 \\
& KD~\shortcite{vanillakd} & 70.67 & 73.08 & 74.92 & 73.33 \\
& DKD~\shortcite{DKD} & - & 74.11  & \textbf{76.24} & \textbf{76.32}  \\
& DIST~\shortcite{DIST} & 69.94 & 73.55  & 75.29 & 75.79   \\
& PCKA(ours) & \textbf{70.97} & \textbf{74.11}  & 75.61 & 74.98  \\ \hline
\end{tabular}}
\end{small}
\label{tab:cifar100patch}
\end{table}

\begin{figure}[ht]
\includegraphics[width=1\linewidth]{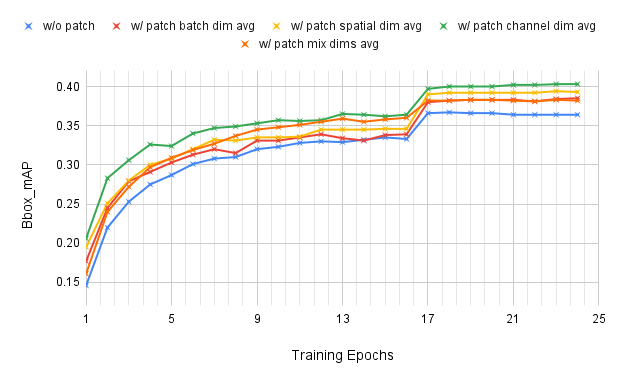}
\caption{The effect of patch on different dimensions. This experiment is conducted on the RetinaNet-X101-RetinaNet-R50 pair}
\label{fig:avg}
\end{figure}

\section{CKA curve in training.}
\label{apd:cka_value}
\begin{figure*}[htbp]
  \centering
  \begin{minipage}[t]{1\linewidth}
      \centering
      \includegraphics[width=1\linewidth]{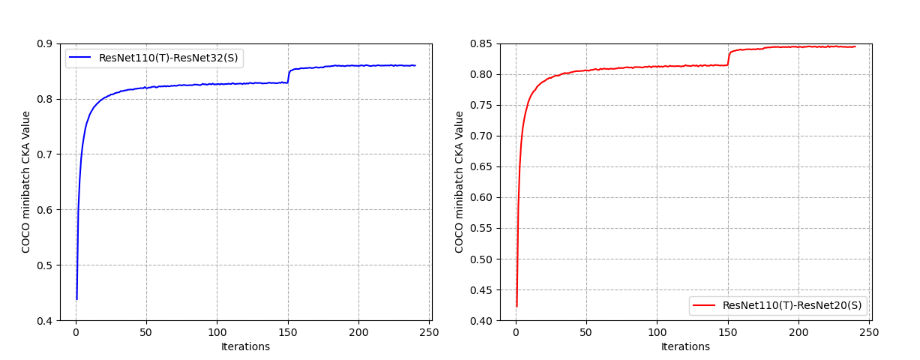}
  \end{minipage}
  \begin{minipage}[t]{1\linewidth}
      \centering
      \includegraphics[width=1\linewidth]{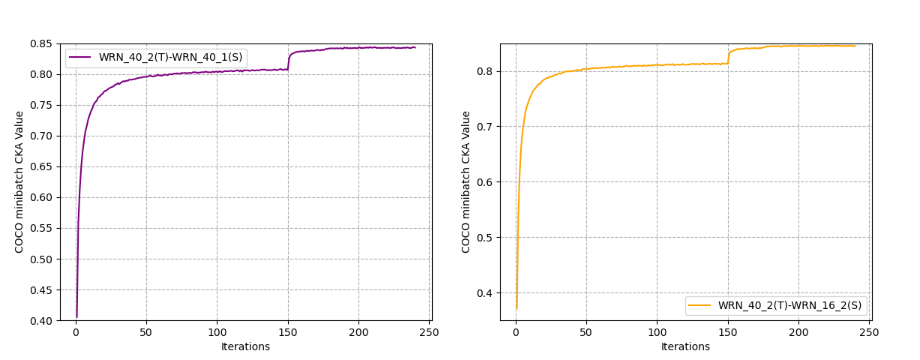}
  \end{minipage}
  \begin{minipage}[t]{1\linewidth}
      \centering
      \includegraphics[width=1\linewidth]{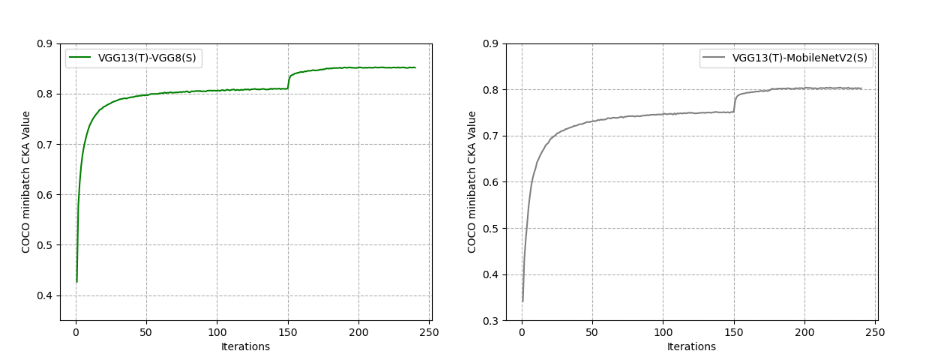}
  \end{minipage}
  \caption{CKA curve in training phase on CIFAR-100. We visualize the CKA similarities in the training for six teacher-student pairs.}
  \label{fig:value}
\end{figure*}

To qualitatively analyze the proposed method, we visualize the CKA similarities in the training phase. As shown in Fig.~\ref{fig:value}, we can find the CKA values increase with training, which demonstrates that our method does narrow the gap of teacher-student models at the logit level.

\begin{table}[!ht]
\caption{Ablation study of Batchsize. Noted that "4 $\times$ 1" means 4 GPUs, each batchsize is 1, totally batchsize is $4\times1=4$}
\resizebox{1.\linewidth}{!}{
\begin{tabular}{l|cccccc}
 \multirow{1}{*}{T$\rightarrow$S} & \multicolumn{6}{c}{RetinaNet-X101$\rightarrow$RetinaNet-R50}  \\ \hline
 & AP & AP$_{50}$ & AP$_{75}$ & AP$_{S}$ & AP$_{M}$ & AP$_{L}$ \\ \hline
Teacher & 41.0 & 60.9 & 44.0 & 23.9 & 45.2 & 54.0  \\
Student & 37.4 & 56.7 & 39.6 & 20.0 & 40.7 & 49.7  \\ \hline
 Batchsize 4$\times$1  & 39.9 & 59.5 & 42.8 & 21.7 & 43.9 & 53.3 \\
 Batchsize 4$\times$2 & \textbf{40.3} & 59.8 & \textbf{43.1} & 22.5 & \textbf{44.2} & 54.4 \\
 Batchsize 4$\times$3 & \textbf{40.3} & \textbf{59.9} & 43.0 & \textbf{23.3} & \textbf{44.2} & \textbf{54.9} \\
 Batchsize 4$\times$4 & 40.0 & 59.8 & 42.6 & 22.9 & 43.8 & 54.1 \\
\end{tabular}}
\label{tab:batchsize}
\end{table}

\begin{table}[!ht]
\caption{Ablation study of number of distillation layers. Noted that "Layer=1" means choosing 1 layer to distill.}
\resizebox{1.\linewidth}{!}{
\begin{tabular}{l|cccccc}
 \multirow{1}{*}{T$\rightarrow$S} & \multicolumn{6}{c}{RetinaNet-X101$\rightarrow$RetinaNet-R50}  \\ \hline
 & AP & AP$_{50}$ & AP$_{75}$ & AP$_{S}$ & AP$_{M}$ & AP$_{L}$ \\ \hline
Teacher & 41.0 & 60.9 & 44.0 & 23.9 & 45.2 & 54.0  \\
Student & 37.4 & 56.7 & 39.6 & 20.0 & 40.7 & 49.7  \\ \hline
 Layer=1  & 39.9 & 59.7 & 42.8 & 21.9 & 43.5 & 54.1 \\
 Layer=2  & 40.2 & \textbf{59.9} & 42.6 & 21.9 & 44.0 & 54.6 \\
 Layer=3  & \textbf{40.3} & \textbf{59.9} & \textbf{43.0} & \textbf{23.3} & \textbf{44.2} & \textbf{54.9} \\ 
\end{tabular}}
\label{tab:distillation_layer}
\end{table}

\section{PCKA in image classification}
We apply PCKA to the classification task, and it also outperforms well on the methods with the same distillation type, as shown in Table~\ref{tab:cifar100patch}. However, PCKA performs very badly on the teacher and student pairs with different architectures. The possible reason is that the cutting activations of different architectures contain more dissimilar and harmful representations, causing difficulty in transferring knowledge to the student.

\begin{figure*}[ht]
\includegraphics[width=1\textwidth]{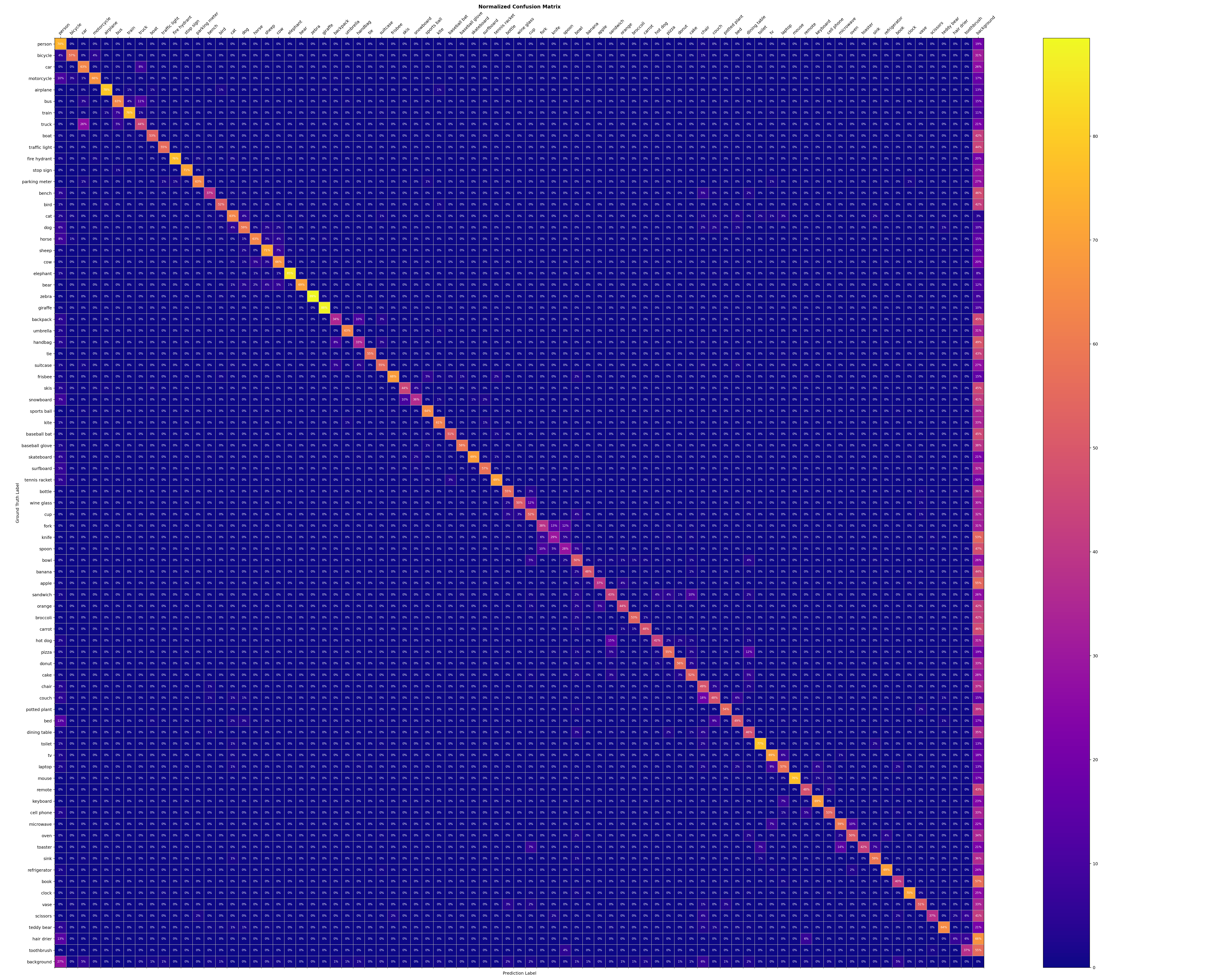}
\caption{The confusion matrix is from the RetinaNet-R50, distilled by PCKA with teacher model RetinaNet-X101, averaging on channel.}
\label{fig:confusion}
\end{figure*}

\begin{figure*}[ht]
\includegraphics[width=1\textwidth]{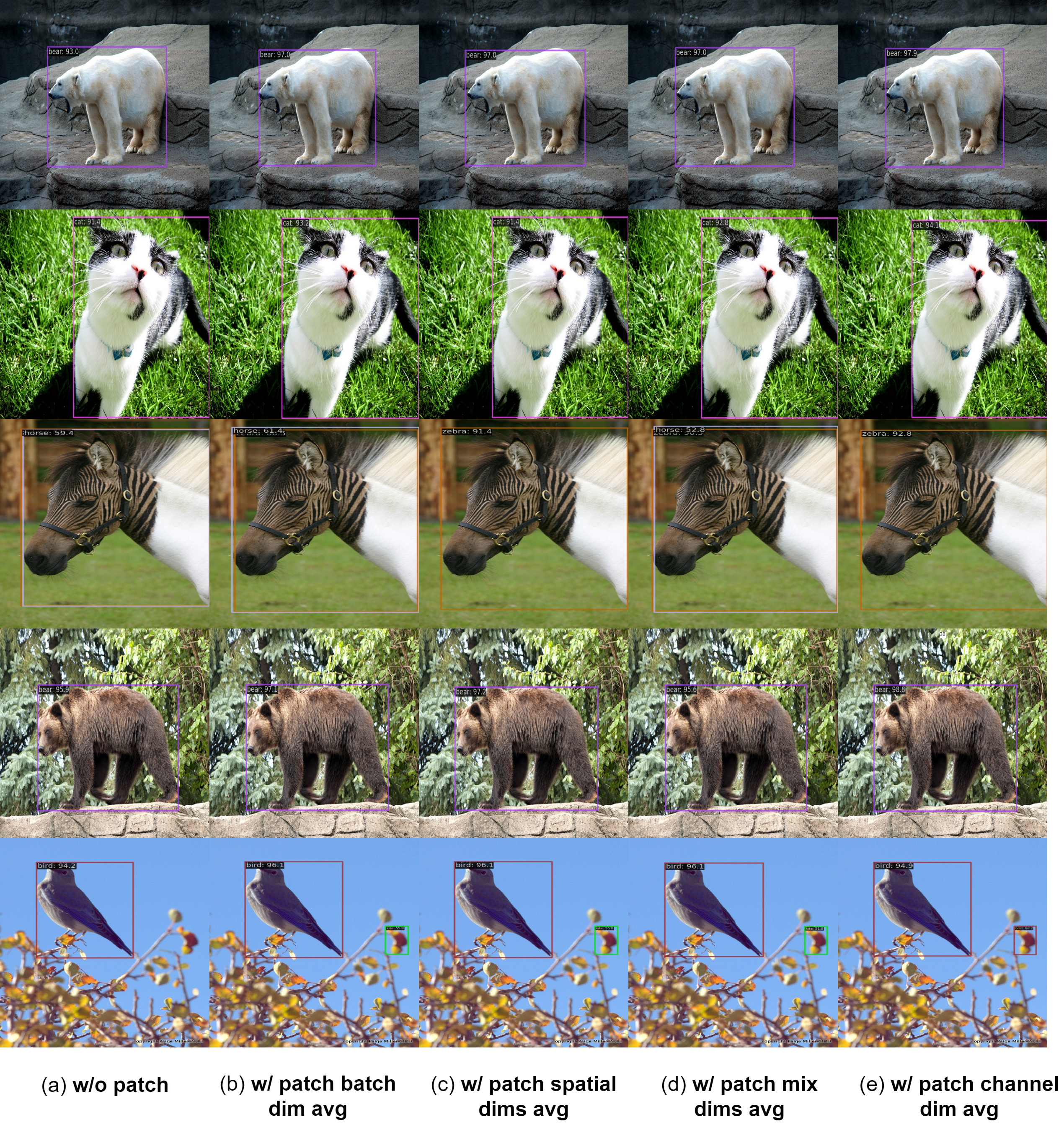}
\caption{Top 5 inference results with RetinaNet-R50, distilled by PCKA without patch or with patch.}
\label{fig:vis}
\end{figure*}

\end{document}